\documentclass{kluwer}

\newcommand{\citet}[1]{\citeauthor{#1} \shortcite{#1}}
\newcommand{\citealp}[1]{\citeauthor{#1}, \citeyear{#1}}
\newcommand{\citep}[1]{\cite{#1}}

\usepackage{amsmath}
\usepackage{amssymb}

\usepackage[OT4,T1]{fontenc}
\usepackage{url}
\usepackage{alltt}
\usepackage{array}
\usepackage{examples}
\newcommand{\okra}[1]{\left( #1 \right)}
\newcommand{\kwad}[1]{\left[ #1 \right]}
\newcommand{\klam}[1]{\left\{ #1 \right\}}
\newcommand{\sred}[1]{\left\langle #1 \right\rangle}

\newcommand{\boole}[1]{{\bf 1}_{\klam{#1}}}

\newcommand{\subst}[1]{\left. #1 \right|}
\newcommand{\card}[1]{\left| #1 \right|}
\newcommand{\argmax}{\mathop{\arg \max}\displaylimits}
\newcommand{\ind}{\,\mathrel{\mkern-5mu\perp\mkern-10mu\perp\mkern-5mu}\,}

\newcommand{\excl}{\times}
\newcommand{\inde}{\perp}
\newcommand{\rarr}{\rightarrow}
\newcommand{\larr}{\leftarrow}
\newcommand{\earr}{\leftrightarrow}

\newcommand{\lemat}[1]{\text{\emph{#1}}}
\newcommand{\znacz}[1]{\text{\textsf{#1}}}

\newcommand{\FRA}{\textbf{F}}
\newcommand{\LIS}{\textbf{L}}
\newcommand{\MAT}{\textbf{M}}
\newcommand{\EXP}{\textbf{E}}

\begin{document}

\begin{opening}         
  \title{Valence extraction using EM selection and co-occurrence
    matrices} 
  \author{{\L}ukasz \surname{D\k{e}bowski}\footnote{%
      The author is presently on leave for Centrum Wiskunde \&
      Informatica, Science Park 123, NL-1098 XG Amsterdam, the
      Netherlands. E: \url{debowski@cwi.nl}, T: +31 20 592 4193,
      F: +31 20 592 4312.}}  
  \runningauthor{{\L}ukasz D\k{e}bowski}
  \runningtitle{Valence extraction using EM selection and
    co-occurrence matrices}
  \institute{Instytut Podstaw Informatyki PAN\\
    J.K. Ordona 21, 01-237 Warszawa, Poland}


\begin{abstract}
  This paper discusses two new procedures for extracting verb valences
  from raw texts, with an application to the Polish language.  The
  first novel technique, the EM selection algorithm, performs
  unsupervised disambiguation of valence frame forests, obtained by
  applying a~non-probabilistic deep grammar parser and some
  post-processing to the text.  The second new idea concerns filtering
  of incorrect frames detected in the parsed text and is motivated by
  an observation that verbs which take similar arguments tend to have
  similar frames. This phenomenon is described in terms of newly
  introduced co-occurrence matrices. Using co-occurrence matrices, we
  split filtering into two steps. The list of valid arguments is first
  determined for each verb, whereas the pattern according to which the
  arguments are combined into frames is computed in the following
  stage. Our best extracted dictionary reaches an $F$-score of $45\%$,
  compared to an $F$-score of $39\%$ for the standard frame-based BHT
  filtering.
\end{abstract}
\end{opening}

\section{Introduction}
\label{secIntro}

The aim of this paper is to explore two new techniques for verb
valence extraction from raw texts, as applied to the Polish language.
The methods are novel compared to the standard framework
\citep{Brent93,Manning93,ErsanCharniak95,BriscoeCarroll97} and
motivated in part by resources available for this language and in part
by certain linguistic observations.

The task of valence extraction for Polish invites novel approaches
indeed.  Although there is no treebank for this language on which
a~probabilistic parser can be trained, a~few interesting resources are
available. Firstly, the non-probabilistic parser \'Swigra
\citep{Wolinski04,Wolinski05} provides an efficient implementation of
the large formal grammar of Polish by \citet{Swidzinski92}.  Secondly,
three detailed valence dictionaries have been compiled by formal
linguists \cite{Polanski80,Swidzinski94,Banko00}. Those dictionaries
are potentially useful as a~gold standard in automatic valence
extraction but two of them, \citeauthor{Polanski80} and
\citeauthor{Banko00}, are printed on paper in several volumes, whereas
\citeauthor{Swidzinski94}'s dictionary, though rather small, is
available electronically. The text file by \citeauthor{Swidzinski94}
lists about 1000 verbal entries whereas 6000 entries can be found in
COMLEX, a~detailed syntactic dictionary of English
\cite{MacleodGrishmanMeyers94}.

The information provided by Polish valence dictionaries is of
comparable complexity to information available in COMLEX. Verbs in the
dictionary entries select for nominal (NP) and prepositional (PP)
phrases in specific morphological cases (7 distinct cases and many more
prepositions).  Valence frames may contain the reflexive marker
\lemat{si\k{e}} and certain adjuncts (e.g., adverbs) but not
necessarily a~subject, which also contributes to the combinatorial
explosion. For instance, \citet{Swidzinski94} provides 329 frame types
for the 201 test verbs described later in Section
\ref{secEvaluation}. The most frequent frame among them,
$\klam{\znacz{np(nom)},\znacz{np(acc)}}$, is valid for 124 test verbs
and there are 183 hapax frames.

Such lack of computational data is a~strong incentive to develop
automatic valence extraction as efficiently as possible.  Thus we have
devised two procedures.  The first one, called the EM selection
algorithm, performs unsupervised selection of alternative valence
frames. These frames were obtained for sentences in a~corpus by
applying the parser \'Swigra and some post-processing.  In this way,
we cope with the lack of a~probabilistic parser and of a~treebank.

The EM selection procedure, to our knowledge described here for the
first time, assumes that the disambiguated alternatives are highly
repeatable atomic entities. The procedure does not rely on what formal
objects the alternatives are but it only takes their frequencies into
account. Thus, the EM selection looks like an interesting baseline
algorithm for many unsupervised disambiguation problems, e.g.\
part-of-speech tagging
\citep{Kupiec92,Merialdo94}. Computationally, the algorithm is far
simpler than the inside-outside algorithm for probabilistic grammars
\citep{ChiGeman98}, which also instantiates the
expectation-maximization scheme and is used for treebank and valence
acquisition \citep{BriscoeCarroll97,CarrollRooth98}.

The second novel technique concerns filtering of incorrect valence
frames detected in the parsed text. Despite a~large number of distinct
frames occurring in the available Polish valence dictionaries, verbs
which take similar arguments tend to have similar frames. This
phenomenon was surveyed in particular by \citet{DebowskiWolinski07}
and their observations are reported here in more detail in Section
\ref{secFormalism}. The cited authors proposed that sets of verbal
frames be described in terms of argument lists, which strongly depend
on a~verb, and pairwise combination rules for arguments, called
co-occurrence matrices, which are largely independent of a~verb.  

In this article, we recall this formalism and propose an analogous
two-stage approach to filtering incorrect frames. The list of
arguments is filtered for each verb initially and then the
co-occurrence matrices are processed. In both steps we use filtering
methods that resemble those used so far for whole frames. We will show
that verbal frames are easier to extract when decomposed into simpler
entities than when treated as atomic objects. The qualitative analysis
of errors is also easier to perform.

Verb valence frames have been learned as atomic entities in all
previous valence extraction experiments (see also:
\citealp{SarkarZeman00}; \citealp{PrzepiorkowskiFast05};
\citealp{PrzepiorkowskiFast05b}; \citealp{ChesleySalmonAlt06})
although recent research exploits certain correlations among the verb
meanings, diathesis, and subcategorization (\citealp{McCarthy01};
\citealp{Korhonen02}, Chapter 4; \citealp{LapataBrew04};
\citealp{SchulteImWalde06}). This line of computational experiment is
more and more inspired by formal research in semantic classes of
arguments, verbs, and frame alternations, cf.\ \citet{Levin93} and
\citet{BakerRuppenhofer02}.  

Our unorthodox less resource- and theory-intensive approach to
decomposing valence frames stems from an independent insight into
their distribution and structure, built on the preliminary valence
extraction experiment for Polish by \citet{PrzepiorkowskiFast05}. In
that experiment, the $F$-score of the automatically extracted
dictionary reached about 40\%, whereas the $F$-score of two
gold-standard dictionaries by \citet{Polanski80} and \citet{Banko00}
compared with each other equalled 65\%. This apparently low agreement
between manually compiled dictionaries and the lack of explicit
information about semantic classes inspired us to seek other patterns
in valence frames and to develop an alternative extraction scheme.

The experiment described in this paper differs from both of the works
by \citeauthor{PrzepiorkowskiFast05} in several aspects. Firstly, we
explore whether it is better to filter frames in two steps or in one
step as done previously. Secondly, we extract all kinds of arguments
occurring in the gold-standard dictionaries, whereas only non-subject
NPs and PPs were considered in the two previous works.  Thirdly, we
compare our extracted dictionaries with three gold-standard
dictionaries simultaneously and investigate types of errors.
Fourthly, we use the \'Swigra parser of \citeauthor{Swidzinski92}'s
grammar and the EM algorithm to parse raw texts, whereas
\citeauthor{PrzepiorkowskiFast05} applied a~very simple regular
grammar of 18 rules. We analyze fewer texts but we analyze them more
thoroughly, which means higher precision but not necessarily lower
recall. The final difference is that our test set covers twice as many
verbs (201 lemmas) as considered by \citeauthor{PrzepiorkowskiFast05}.

The frame-based binomial hypothesis test (BHT, \citealp{Brent93}) is
assumed in this work as a~baseline against which our new ideas of
filtering are compared, since it gave the best results according to
\citet{PrzepiorkowskiFast05b}. The authors reapplied several known
frame filtering methods: the BHT, the log-likelihood ratio test (LLR)
\citep{Gorrell99,SarkarZeman00}, and the maximum likelihood threshold
(MLE) \citep{Korhonen02}. Applying the one-stage BHT to our data, we
obtain $26\%$ recall and $75\%$ precision ($F=39\%$). To compare, the
dictionary obtained by applying the novel two-stage filtering of
frames to the same counts of parses exhibits $32\%$ recall and $60\%$
precision ($F=42\%$). The set-theoretic union of both dictionaries
combines their strengths and features $F=45\%$. These statistics relate
to extracting whole frames, whereas \citeauthor{PrzepiorkowskiFast05}
obtained similar values for the simpler task of extracting only NPs
and PPs. We find our results to be an encouraging signal that
similarities of frame valence frame sets should be exploited across
different verbs as much as possible, and also in an algorithmic
way. The method introduced here allows various extensions and
modifications.

The rest of this article describes our experiment in more detail. In
Section \ref{secFormalism}, a~brief introduction to co-occurrence
matrices is provided; Section \ref{secExtraction} presents the verb
valence extraction procedure; the obtained dictionary is analyzed in
Section \ref{secEvaluation}; Section \ref{secConclusion} contains the
conclusion. Three appendices follow the article. Appendix
\ref{appMatrix} gives additional details for the co-occurrence matrix
formalism; Appendix \ref{appParsing} describes the initial corpus
parsing; Appendix \ref{appEM} introduces the EM selection algorithm.

\section{The formalism of co-occurrence matrices}
\label{secFormalism}

Let us introduce the new description of valence frames which is
applied to valence extraction in this paper. To begin with
a~more usual formal concept, consider a~prototypical entry from our
gold-standard valence dictionary. It consists of the set of valence
frames
\begin{align}
\label{Fprzylapac}
 \FRA(\lemat{przy{\l}apa\'c})=
  \klam{
    \begin{array}[c]{l}
    \klam{\znacz{np(nom)},\znacz{np(acc)}},\\
    \klam{\znacz{np(nom)},\znacz{np(acc)},\znacz{na+np(loc)}},\\
    \klam{\znacz{np(nom)},\znacz{sie},\znacz{na+np(loc)}}\\
    \end{array}
  } 
\end{align}
for the verb \lemat{przy{\l}apa\'c} (=~\lemat{to catch somebody
  red-handed}). The symbol \znacz{sie} denotes the reflexive marker
\lemat{si\k{e}} and \znacz{na+np(loc)} is a~prepositional phrase with
preposition \lemat{na} (=~\lemat{on)}, which requires a~nominal phrase
in the locative case. The notations for cases are as in the IPI PAN
Corpus tagset: \znacz{nom}(inative), \znacz{gen}(itive),
\znacz{dat}(ive), \znacz{acc}(usative), \znacz{inst}(rumental),
\znacz{loc}(ative), and \znacz{voc}(ative), cf.\
\citet{PrzepiorkowskiWolinski03} or \url{http://korpus.pl/}. For
simplicity, it is assumed that no argument type can be repeated in
a~single valence frame.  This restriction can be overcome by assigning
unique identifiers to repetitions.

There are two subtleties which concern our implementation of notation
(\ref{Fprzylapac}) and are worth exposing to avoid possible confusion
later:
\begin{enumerate}
\item[(i)] We treat the reflexive marker \lemat{si\k{e}} as an
  ordinary verb argument rather than as a~part of the verb lemma. The
  frames for a~verb without \lemat{si\k{e}} are merged with the frames
  of its possible counterpart with \lemat{si\k{e}} into one entry,
  unlike the traditional linguistic analysis applied in Polish valence
  dictionaries. This affects all our counts of verb entries in the
  following work. However, we do not combine entries for corresponding
  perfective and imperfective verbs, which often take the same frames
  and occur in almost complementary pairs, cf.\ \citet{Mlynarczyk04}.
\item[(ii)] A~valence frame may lack the subject \znacz{np(nom)}.
  According to the analysis applied in Polish dictionaries, this lack
  is a~counterpart of the English expletive \lemat{it} and it differs
  syntactically to the dropped subject (denoted always as
  \znacz{np(nom)} in the valence frame for a~sentence). If a~sentence
  lacks an overt subject, such a~subject can or cannot be inserted
  depending on the verb. Certain verbs do not subcategorize for
  subject at all, e.g.\ \lemat{trzeba} (=~\lemat{should}) or
  \lemat{brakowa\'c} (=~\lemat{lack}). Several other verbs often occur
  without the subject but allow it in certain uses, such as
  \lemat{pada\'c} (=~\lemat{fall/rain}). The valences of the second
  class of verbs are particularly hard to extract automatically since
  Polish is a~pro-drop language.
\end{enumerate}
Summarising our remarks, there are many specific verbs such that
\znacz{sie} or \znacz{np(nom)} (a) must appear in all their frames,
(b) cannot appear in any frames, or (c) may be present or omitted,
affecting the occurrence of other arguments. Similar interactions
involving the reflexive marker and the subject have been studied in
valence acquisition for other languages
\citep{MayolBoledaBadia05,SurdeanuMoranteMarquez08}.

\citet{DebowskiWolinski07} proposed an approximate description of
complex interactions within the frame set $\FRA(v)$ in terms of three
simpler objects: the set of possible arguments $\LIS(v)$, the set of
required arguments $\EXP(v)\subset \LIS(v)$, and the argument
co-occurrence matrix $\MAT(v):\LIS(v)\times \LIS(v)\rarr
\klam{\larr,\rarr,\earr,\excl,\inde}$. The definitions of the first
two objects correspond to the following naming convention.  An
argument is possible for $v$ if it appears in at least one frame and
it is called required for $v$ if it occurs in all frames.  Thus we
have
\begin{align}
\label{LIS}
  \LIS(v)&:=\bigcup_{f\in \FRA(v)}\, f,
&
  \EXP(v)&:=\bigcap_{f\in \FRA(v)}\, f.
\end{align}
For instance, 
\begin{align*}
  \LIS(\lemat{przy{\l}apa\'c})&=
  \klam{\znacz{np(nom)},\znacz{np(acc)},\znacz{sie},\znacz{na+np(loc)}}  
  ,
  \\
  \EXP(\lemat{przy{\l}apa\'c})&= 
  \klam{\znacz{np(nom)}}
  .
\end{align*}

To define the co-occurrence matrix, let us denote the set of verb
frames which contain an argument type $a$ as $\sred{a}:=\klam{f\in
  \FRA(v)\mid a\in f}$.  Next, we will introduce five implicitly
verb-dependent relations:
\begin{align*}
a\excl b &\iff \sred{a}\cap \sred{b}=\emptyset
&&
\text{($a$ excludes $b$)},
\\
a\earr b &\iff \sred{a}=\sred{b}
&&
\text{($a$ and $b$ co-occur)},
\\
a\rarr b &\iff \sred{a}= \sred{a}\cap \sred{b}\not=\sred{b}
&&
\text{($a$ implies $b$)},
\\
a\larr b &\iff \sred{a}\not = \sred{a}\cap \sred{b}=\sred{b}
&&
\text{($b$ implies $a$)},
\\
a\inde b &\iff  \sred{a}\cap \sred{b}\not\in
\klam{\sred{a}, \sred{b}, \emptyset}
&&    
\text{($a$ and $b$ are independent)}.
\end{align*}
Then the cells of matrix $\MAT(v)$
are defined via the equivalence
\begin{align}
\label{MAT}
\MAT(v)_{ab}:=\text{R} \iff a\, \text{R}\, b 
\end{align}
for the verb arguments $a,b\in \LIS(v)$.  The symbol $\inde$ that denotes
``formal'' independence was chosen intentionally to resemble the symbol
$\ind$, which is usually applied to denote probabilistic independence.

For the discussed example we obtain:
\begin{align*}
    \begin{array}[b]{l|cccc}
\MAT(\lemat{przy{\l}apa\'c})
&\znacz{np(nom)}  &\znacz{np(acc)} &\znacz{sie} &\znacz{na+np(loc)} \\
\hline
\znacz{np(nom)}   &\earr &\larr &\larr &\larr \\
\znacz{np(acc)}   &\rarr &\earr &\excl &\inde \\
\znacz{sie}       &\rarr &\excl &\earr &\rarr \\
\znacz{na+np(loc)}&\rarr &\inde &\larr &\earr \\
  \end{array}
  .
\end{align*}

This unconventional approach to describing verb valences appears quite
robust.  For example, consider an observed agreement score (cf.\
\citealp{ArtsteinPoesio08}) of the co-occurrence matrix cells taken for
the triples $(a,b,v)$ appearing simultaneously in two compared
dictionaries.  Formally this agreement score equals
\begin{align}
  \label{AgrRate}
  A_o:=\frac{
    \card{\klam{
      (a,b,v)\in T \mid \MAT_1(v)_{ab}=\MAT_2(v)_{ab}}
    }
  }{
    \card{T}
  }
  ,
\end{align}
where $\klam{\MAT_i(v)\mid v\in V_i}$, $i=1,2$, are the two compared
collections of co-occurrence matrices and
$$T=\klam{(a,b,v)\mid v\in V_1\cap V_2,\, a,b\in
  \LIS_1(v)\cap\LIS_2(v)}$$ is the appropriate subset of triples
$(a,b,v)$. The agreement scores (\ref{AgrRate}) for the dictionaries
of \citeauthor{Polanski80}, \citeauthor{Swidzinski94}, and
\citeauthor{Banko00} range from $86\%$ to $89\%$, cf.\
\citet{DebowskiWolinski07}. 


\citeauthor{DebowskiWolinski07} noticed also that the values of the
matrix cells $\MAT(v)_{ab}$ for fixed arguments $a$ and $b$ tend not
to depend on the verb $v$. The latter fact appears favourable for
automatic valence extraction. We may learn objects $\LIS(v)$,
$\EXP(v)$, and $\MAT(v)$ separately with much higher accuracy and
restore the set of frames $\FRA(v)$ from these by
approximation. For example, consider the maximal set
$\bar\FRA(v)\subset 2^{\LIS(v)}$ of frames that contain all required
arguments in $\EXP(v)$ and induce the co-occurrence matrix $\MAT(v)$.
Precisely,
\begin{align}
\label{OArg}
\bar\FRA(v):=
\klam{f\in 2^{\LIS(v)}\middle|
    \begin{array}[c]{l}
    \forall_{a\in \EXP(v)}\, a\in f
    ,\\
    \forall_{a,b\in \LIS(v)}\, \phi(f,\MAT(v),a,b)      
    \end{array}
  }
   ,
\end{align}
where
\begin{align*}
  \phi(f,\mu,a,b)
  :=
  \begin{cases}
    \neg(a\in f \land b\in f), & \mu_{ab}=\excl, \\
    a\in f \iff b\in f, & \mu_{ab}=\,\earr, \\
    a\in f \implies b\in f, & \mu_{ab}=\,\rarr, \\
    a\in f \impliedby b\in f, & \mu_{ab}=\,\larr, \\
    \text{true}, & \mu_{ab}=\,\inde.
  \end{cases}
\end{align*}

It is easy to see that $\bar\FRA(v)\supset\FRA(v)$. We have
$\bar\FRA(v)\not=\FRA(v)$ for some verbs, as shown in Subsection
\ref{ssecEvalCoo}.  In our application, however, the number of frames
introduced by using $\bar\FRA(v)$ rather than $\FRA(v)$ is small, see
the last paragraph of Subsection \ref{ssecEvalCoo}.  $\bar\FRA(v)$ may
be used conveniently also for syntactic parsing of
sentences. Typically, a~grammar parser checks whether a~hypothetical
frame $f$ of the parsed sentence belongs to the set $\FRA(v)$, defined
by a~valence dictionary linked to the parser. If $\bar\FRA(v)$ rather
than $\FRA(v)$ is used for parsing, which enlarges the set of accepted
sentences, then there is no need to compute $\bar\FRA(v)$ in order to
check whether $f\in\bar\FRA(v)$. The parser can use a~valence
dictionary which is stored just as the triple
$(\LIS(v),\EXP(v),\MAT(v))$.  In our application, however, the
reconstructed set $\bar\FRA(v)$ is needed explicitly for dictionary
evaluation. Thus we provide an efficient procedure to compute
$\bar\FRA(v)$ in Appendix \ref{appMatrix}.

\section{The adjusted extraction procedure}
\label{secExtraction}

\subsection{Overview}
\label{subsecOverview}

Our valence extraction procedure consists of four distinct
subtasks. 

\medskip \textbf{Deep non-probabilistic parsing of corpus data:}
The first task was parsing a~part of the IPI PAN Corpus of Polish to
obtain a~bank of reduced parse forests, which represent alternative
valence frames for elementary clauses suggested by
\citeauthor{Swidzinski92}'s grammar.  The details of this procedure
are described in Appendix \ref{appParsing}.

The obtained bank included $510\,743$ clauses which were decorated
with reduced parse forests like the following two examples (correct
reduced parses marked with a~'+'):
\begin{alltt}
'Kto zast\k{a}pi piekarza?'
  (= 'Who will replace the baker?')
+zast\k{a}pi\'c :np:acc: :np:nom:
zast\k{a}pi\'c :np:gen: :np:nom:
'Nie p{\l}aka{\l} na podium.'
  (= 'He did not cry on the podium.')
p{\l}aka\'c :np:nom: :prepnp:na:acc:
+p{\l}aka\'c :np:nom: :prepnp:na:loc:
\end{alltt}
Reduced parses are intended to be the alternative valence frames for
a~clause plus the lemma of the verb. In contrast to full parses of
sentences, reduced parses are highly repeatable in the corpus
data. Thus, unsupervised learning can be used to find approximate
counts of correct parses in the reduced parse forests and to select
the best description for a~given sentence on the basis of its
frequency in the whole bank.

\medskip \textbf{EM disambiguation of reduced parse forests:} In the
second subtask, the reduced parse forests in the bank were indeed
disambiguated to single valence frames per clause.  It is a~standard
approach to disambiguate full parse forests with a~probabilistic
context-free grammar (PCFG). However, reformulating
\citeauthor{Swidzinski92}'s metamorphosis grammar as a~pure CFG and
the subsequent unsupervised (for the lack of a~treebank) PCFG training
would take too much work for our purposes. Thus we have disambiguated
reduced parse forests by means of the EM selection algorithm
introduced in Appendix \ref{appEM}.  Let $A_i$ be the set of reduced
parse trees for the $i$-th sentence in the bank, $i=1,2,...,M$.  We
set the initial $p_j^{(1)}=1$ and applied the iteration
(\ref{EM1a})--(\ref{EM2a}) from Appendix \ref{appEM} until
$n=10$. Then one of the shortest parses with the largest conditional
probability $p_{ji}^{(n)}$ was sampled at random.

Just to investigate the quality of this disambiguation, we prepared
a~test set of 190 sentences with the correct reduced parses indicated
manually. Since the output of our disambiguation procedure is
stochastic and the test set was small, we performed 500 Monte Carlo
simulations on the whole test set.  Our procedure chose the correct
reduced parse for $72.6\%$ sentences on average.  Increasing the
number of the EM iterations to $n=20$ did not improve this result. As
a comparison, sampling simply a~parse $j$ with the largest
$p_{ji}^{(n)}$ yielded an accuracy of $72.4\%$, sampling a~parse with
the minimal length was accurate in $57.5\%$ cases, whereas blind
sampling (assuming equidistribution) achieved $46.9\%$. The difference
between $72.6\%$ and $72.4\%$ is not significant but, given that it
does not spoil our results, we prefer using shorter parses.


\medskip \textbf{Computing the preliminary dictionary from parses:}
Once the reduced parse forests in the bank had been disambiguated,
a~frequency table of the disambiguated reduced parses was
computed. This will be referred to as the preliminary valence
dictionary. The entries in this dictionary looked like this:
\begin{alltt}
'przy{\l}apa\'c' => \{
    'np(acc),np(gen),np(nom)' => 1, 
+   'na+np(loc),np(nom),sie' => 1,
    'na+np(loc),np(gen),np(nom)' => 1,
+   'np(acc),np(nom)' => 4,
    'adv,np(nom)' => 1,
+   'na+np(loc),np(acc),np(nom)' => 3
\}
\end{alltt}
The numbers are the obtained reduced parse frequencies, whereas the
correct valence frames are marked with a~`+', cf.\
(\ref{Fprzylapac}). Notice that the counts for each parse are low. We
chose a~low frequency verb for this example to make it short.
Another natural method to obtain a~preliminary dictionary was to use
$Mp_j^{(n)}$ coefficients as the frequencies of frames. This method
yields final results that are $1\%$ worse than for the dictionary based
on the frequency table.

\medskip \textbf{Filtering of the preliminary dictionary:} The
preliminary dictionary contains many incorrect frames, which are due
to parsing or disambiguation errors. In the last subtask, we filtered
this dictionary using supervised learning, as done commonly in related
work.

For example, the BHT filtering by \citet{Brent93} is as follows. Let
$c(v,f)$ be the count of reduced parses in the preliminary dictionary
that contain both verb $v$ and valence frame $f$. Denote the frequency
of verb $v$ as $c(v)=\sum_f c(v,f)$. Frame $f$ is retained in the set
of valence frames $\FRA(v)$ if and only if
\begin{align}
  \label{FraThreshold}
\sum_{n=c(v,f)}^{c(v)} \binom{c(v)}{n} p_{f}^n (1-p_{f})^{c(v)-n}
\le \alpha
,
\end{align}
where $\alpha=0.05$ is the usual significance level and $p_{f}$ is
a~frequency threshold. The parameter $p_f$ is selected as a~value for
which the classification rule (\ref{FraThreshold}) yields the minimal
error rate against the training dictionary. In the idealized language
of statistical hypothesis testing, $p_{f}$ equals the empirical
relative frequency of frame $f$ for the verbs that \emph{do not
  select} for $f$ according to the ideal dictionary.

We have used the BHT as the baseline, against which we have tested
a~new procedure of frame filtering. The new procedure applied the
co-occurrence matrices presented in Section \ref{secFormalism}. It was
as follows:
\begin{enumerate}
\item Compute $\LIS(v)$ and $\EXP(v)$ via Equation (\ref{LIS}) 
  from the sets of valence frames $\FRA(v)$ given by the
  preliminary dictionary.
\item Correct  $\LIS(v)$ and $\EXP(v)$ using the training dictionary.
\item Reconstruct $\FRA(v)$ given the new $\LIS(v)$ and
  $\EXP(v)$. This reconstruction is defined 
  as the substitution $\FRA(v)\gets 
  \klam{ (f\cup \EXP(v))\cap \LIS(v) \mid f\in\FRA(v) } $.
\item Compute $\MAT(v)$ from $\FRA(v)$ via Equation (\ref{MAT}).
\item Correct $\MAT(v)$ using the training dictionary.
\item Reconstruct $\FRA(v)$ given the new $\MAT(v)$.  This
  reconstruction consists of substitution $\FRA(v)\gets \bar\FRA(v)$,
  where $\bar\FRA(v)$ is defined in Equation (\ref{OArg}) and computed
  via the procedure described in Appendix \ref{appMatrix}.
\item Output $\FRA(v)$ as the valence of verb $v$.
\end{enumerate}
Steps 2.\ and 5.\ are described in Subsections \ref{subsecArguments}
and \ref{subsecMatrix} respectively.

In our experiment, the training dictionary consisted of valence frames
for 832 verbs from the dictionary of \citet{Swidzinski94}. It
contained all verbs in \citeauthor{Swidzinski94}'s dictionary except
those included in the test set introduced in Section
\ref{secEvaluation}.

\subsection{Filtering of the argument sets}
\label{subsecArguments}

For simplicity of computation, the correction of argument sets
$\LIS(v)$ and $\EXP(v)$ was done by setting thresholds for the
frequency of arguments as in the maximum likelihood thresholding test
for frames (MLE) proposed by \citet{Korhonen02}.  Thus a~possible
argument $a$ for verb $v$ was retained if it accounted for a certain
proportion of the verb's frames in the corpus.  Namely, $a$ was kept
in $\LIS(v)$ if and only if
\begin{align}
  \label{ArgThreshold}
  c(v,a)\ge p_{a} c(v) + 1
  ,
\end{align}
where $c(v)$ is the frequency of reduced parses in the preliminary
dictionary that contain $v$, as in (\ref{FraThreshold}), and $c(v,a)$
is the frequency of parses that contain both $v$ and $a$. Parameter
$p_{a}$ was evaluated as dependent on the argument but independent of
the verb. The optimal $p_{a}$ was selected as a~value for which the
classification rule (\ref{ArgThreshold}) yielded the minimal error
rate against the training dictionary.

The difference between the BHT and the MLE is negligible if the count
of the verb $c(v)$ and the frequency threshold $p_{a}$ are big
enough. This condition is not always satisfied in our application but
we preferred MLE for its computational simplicity and its lack of need
to choose an appropriate significance level $\alpha$. In a~preceding
subexperiment, we had also tried out the more general model $c(v,a)\ge
p_{a} c(v) + t_{a}$ instead of (\ref{ArgThreshold}), where $t_{a}$ was
left to vary. Since $t_{a}=1$ was learned for the vast majority of
$a$'s then we set constant $t_{a}=1$ for all verb arguments later.

Since the same error rate could be obtained for many different values
of $p_{a}$, we applied a~discrete minimization procedure to avoid
overtraining and excessive searching.  Firstly, the resolution level
$N:=10$ was initialized. In the following loop, we checked the error
rate for each $p_{a}:= n/N$, $n=0,1,...,N$. The number of distinct
$p_{a}$'s yielding the minimal error rate was determined and called
the degeneration $D(N)$. For $D(N)<10$, the loop was repeated with
$N:= 10\, N$. In the other case, the optimal $p_{a}$ was returned as
the median of the $D(N)$ distinct values that allowed the minimal
error rate. Selecting the median was inspired by the maximum-margin
hyperplanes used in support vector machines to minimize overtraining
\citep{Vapnik95}.

Similar supervised learning was used to determine whether a~given
argument is strictly compulsory for a~verb.  By symmetry, an argument
$a$ that was found possible with verb $v$ was considered as required
unless it was rare enough. Namely, $a\in\LIS(v)$ was included in the
new $\EXP(v)$ unless
\begin{align}
  \label{ArgThreshold2}
  c(v)-c(v,a)\ge p_{\neg a} c(v) + 1
  ,
\end{align}
where $p_{\neg a}$ was another parameter, estimated analogously to $p_{a}$.

\subsection{Correction of the co-occurrence matrices}
\label{subsecMatrix}

Once we had corrected the argument sets in the preliminary dictionary,
the respective co-occurrence matrices still contained some errors when
compared with the training dictionary.  However, the number of those
errors was relatively small and it was not so trivial to propose an
efficient scheme for their correction.

A~possible approach to such correction is to develop statistical tests
with clear null hypotheses that would detect structural zeroes in
contingency tables
\begin{align*}
  \begin{array}[c]{|c|c|c|}
    \hline
    & 
    a\not\in f 
    & a\in f 
    \\
    \hline
     b\not\in f
    & N - N_a - N_b + N_{ab}
    & N_a - N_{ab}
    \\
    \hline
    b\in f 
    & N_b - N_{ab} 
    & N_{ab} 
    \\
    \hline
  \end{array}
  \,\, ,
\end{align*}
where $N=\card{\FRA(v)}$, $N_a=\card{\sred{a}}$,
$N_b=\card{\sred{b}}$, and $N_{ab}=\card{\sred{a}\cap \sred{b}}$ are
appropriate counts of frames. Relations $\larr$, $\rarr$, $\earr$, and
$\excl$ correspond to particular configurations of structural zeroes
in these tables. 

Constructing structural zero detection tests appeared to be difficult
under the common-sense requirement that the application of these tests
cannot diminish the agreement score (\ref{AgrRate}) between the
corrected dictionary and the training dictionary.  We have
experimented with several such schemes but they did not pass the
aforementioned criterion empirically. Eventually, we have discovered
successful correction methods which rely on the fact that values of
matrix cells for fixed arguments tend not to depend on a verb, see
Section \ref{secFormalism}.

In this paper we compare three such correction methods. Let us
denote the value of a~cell $\MAT(v)_{ab}$ after Step 4 as $\text{S}$.
On the other hand, let $\text{R}$ be the most frequent relation for
arguments $a$ and $b$ given by the training dictionary across
different verbs. We considered the following correction schemes:
\begin{enumerate}
\item[(A)] $\MAT(v)_{ab}$ is left unchanged (the baseline):
  $\MAT(v)_{ab}\gets \text{S}$.
\item[(B)] $\MAT(v)_{ab}$ becomes verb-independent: $\MAT(v)_{ab}\gets \text{R}$.
\item[(C)] We use the most prevalent value only if there is enough
  evidence for a~verb-independent interaction:
  \begin{align}
  \label{MatThreshold}
  \MAT(v)_{ab}\gets
  \begin{cases}
    \text{R},  
    & C(a\, \text{R}\, b)\ge 
    p_{\text{S}\Rightarrow \text{R}} C(a,b) + t_{\text{S}\Rightarrow \text{R}}, 
    \\
    \text{S}, 
    & \text{else},
  \end{cases}
\end{align}
where $C(a\, \text{R}\, b)$ is the number of verbs for which $a\,
\text{R}\, b$ is satisfied and $C(a,b)$ is the number of verbs that
take both $a$ and $b$; both numbers relate to the training
dictionary. Coefficients $p_{\text{S}\Rightarrow \text{R}}$ and
$t_{\text{S}\Rightarrow \text{R}}$ are selected as the values for
which rule (\ref{MatThreshold}) returns the maximal agreement score
(\ref{AgrRate}) against the training dictionary.
\end{enumerate}

There were only a~few relation pairs $\text{S}\Rightarrow \text{R}$
for which method (C) performed substitutions $\MAT(v)_{ab}\gets
\text{R}$ when applied to our data.  These were:
$\larr$$\Rightarrow$$\excl$, $\rarr$$\Rightarrow$$\excl$,
$\inde$$\Rightarrow$$\larr$, $\inde\Rightarrow \rarr$, and
$\inde$$\Rightarrow$$\excl$.  Unlike the case of argument filtering,
the optimal $t_{\text{S}\Rightarrow\text{R}}$ was equal to $1$ only
for one relation pair, namely $\inde$$\Rightarrow$$\excl$.  The
evaluation of methods (A), (B) and (C) against an appropriate test set
is presented in Section \ref{ssecEvalCoo}.

\section{Evaluation of the dictionary}
\label{secEvaluation}

\subsection{Overview}
\label{subsecOverviewII}

Having applied the procedures described in Section
\ref{secExtraction}, we obtained an automatically extracted valence
dictionary that included 5443 verb entries after Step 6, which is five
times more than in \citet{Swidzinski94}.  As mentioned in the previous
section, all parameters were trained on frame sets provided by
\citet{Swidzinski94} for 832 verbs. In contrast, the valence frames in
our test set were simultaneously given by \citet{Swidzinski94},
\citet{Banko00}, and \citet{Polanski80} for 201 verbs different from
the training verbs. Except for 5 verbs missing in
\citeauthor{Polanski80} and one missing in \citeauthor{Banko00}, each
verb in the test set was described by all dictionaries and we kept
track of which dictionary contributed which frame.

We preferred to compare the automatically extracted dictionary with
three reference dictionaries at once to sort out possible mistakes in
them.  In particular, the majority voting (MV) of the three
dictionaries was also considered. The verbs for the test set were
selected by hand for the following reasons: Firstly, each reference
dictionary contained a~different set of verbs in its full
version. Secondly, entries from the dictionaries by
\citeauthor{Banko00} and \citeauthor{Polanski80} had to be typed into
the computer manually and interpreted by an expert since these authors
often described arguments abstractly, like the ``adverbial of
time/direction/cause/degree'', rather than as NPs, PPs or
adverbs. Thirdly, verbs taking rare arguments were intentionally
overrepresented in our test set. Although we could not enlarge or
alter the test set easily to perform reasonable $n$-fold
cross-validation, the variation of scores can be seen by comparing
different automatically extracted dictionaries with different
gold-standard dictionaries. We find this more informative for future
research than the standard cross-validation.

The evaluation is divided into three parts.  We analyze some specific
errors of our two-stage approach, each stage assessed separately.  In
the following, we relate our results to previous research.

\subsection{Analysis of the argument filtering}
\label{subsecTest}

\begin{table*}

\begin{tabular}{lcccccc}
\hline
POSSIBLE       &$p_{a}$     &P      &GSP      &FN     &FP     &E\\
\hline
np(nom) &0.06   &199    &201    &2      &0      &2\\
np(acc) &0.08   &126    &142    &25     &9      &34\\
sie     &0.08   &71     &96     &29     &4      &33\\
np(dat) &0.02   &65     &80     &26     &11     &37\\
np(inst)        &0.04   &39     &61     &31     &9      &40\\
ZE      &0.13   &26     &54     &30     &2      &32\\
adv     &0.18   &56     &46     &23     &33     &56\\
do+np(gen)      &0.07   &25     &46     &25     &4      &29\\
na+np(acc)      &0.06   &17     &41     &25     &1      &26\\
PZ      &0.06   &3      &31     &28     &0      &28\\
w+np(loc)       &0.34   &1      &30     &30     &1      &31\\
z+np(inst)      &0.08   &8      &28     &20     &0      &20\\
BY      &0.14   &4      &28     &26     &2      &28\\
inf     &0.1    &14     &27     &13     &0      &13\\
np(gen) &0.31   &8      &24     &17     &1      &18\\
z+np(gen)       &0.08   &7      &23     &19     &3      &22\\
w+np(acc)       &0.06   &8      &19     &14     &3      &17\\
o+np(loc)       &0.03   &11     &19     &8      &0      &8\\
za+np(acc)      &0.03   &3      &17     &15     &1      &16\\
od+np(gen)      &0.1    &2      &17     &15     &0      &15\\
o+np(acc)       &0.01   &13     &16     &6      &3      &9\\
adj(nom)        &0.77   &1      &3      &2      &0      &2\\
\hline
NOT REQUIRED    &$p_{\neg a}$     &P      &GSP      &FN     &FP     &E\\
\hline
np(nom) &0.54   &3      &19     &19     &3      &22\\
np(acc) &0.24   &174    &174    &10     &10     &20\\
sie     &0.12   &186    &188    &5      &3      &8\\
do+np(gen)      &0.04   &201    &199    &0      &2      &2\\
inf     &0.13   &199    &199    &0      &0      &0\\
np(dat) &0.02   &201    &199    &0      &2      &2\\
\hline
\end{tabular}

\caption{
The evaluation of argument filtering.
} 
  \label{tabArg}
\end{table*}

Table \ref{tabArg} presents the results for parameters $p_{a}$ and
$p_{\neg a}$ tested solely on \citet{Swidzinski94} for the 201 test
verbs.  The notations in the column titles are: P -- the number of
positive outcomes in the automatically extracted dictionary after Step
3 of dictionary filtering (one outcome is one verb taking the
argument), GSP -- the number of gold-standard positive outcomes in
\citeauthor{Swidzinski94} ($\text{GSP}=\text{P}-\text{FP}+\text{FN}$),
FN -- the number of false negatives, FP -- the number of false
positives, and E -- the number of errors
($\text{E}=\text{FN}+\text{FP}$). We have
$0\le\text{FN},\text{FP}\le\text{GSP},\text{P},\text{E}\le 201$.  The
notations for certain arguments in the table rows are: \znacz{sie} --
the reflexive marker \lemat{si\k{e}}, \znacz{x+np(y)} -- the
prepositional phrase introduced by preposition \znacz{x} requiring
a~noun in case \znacz{y}, \znacz{ZE} -- the clause introduced by
\lemat{\.ze} (=~\lemat{that}), \znacz{PZ} -- the clause introduced by
\lemat{czy} (=~\lemat{whether}), and \znacz{BY} -- the clause
introduced by \lemat{\.zeby} (=~\lemat{so as to}).

Although the overall precision of single argument extraction is high
(it reaches $89\%$, see the (verb,~argument) scores in Table
\ref{tabEval} below), all numerical values for this task depend
heavily on the type of extracted argument. The case of frequency
thresholds $p_{a}$, being in the range of $0.02$--$0.77$, is notable.
These thresholds are higher for arguments that can be used as NP
modifiers, e.g.\ \znacz{adj(nom)} and \znacz{np(gen)}, or verbal
adjuncts, e.g.\ \znacz{adv} and \znacz{w+np(loc)}. In general, the
errors concentrate on low-frequency arguments. That occurs probably
because the frequency of tokens coming from parsing errors does not
depend systematically on the argument type. Thus this frequency
dominates the frequency of tokens coming from well parsed sentences
for low-frequency types.  Except for the extraction of a~direct object
\znacz{np(acc)} and adverbial phrase \znacz{adv}, gold-standard
positive outcomes (GSP) outnumber the positive ones (P).  Put
differently, false positives (FP) are fewer than false negatives
(FN)---although the learning objective was set to minimize the error
rate ($\text{E}=\text{FP}+\text{FN}$). The same phenomenon appears in
\citet{Brent93}.

We have also noticed that the extracted valences are better for less
frequent verbs. We can see several reasons for this.  Firstly, there
are more types of infrequent verbs than of frequent ones, so thresholds
$p_{a}$ get more adjusted to the behaviour of less frequent verbs.
Secondly, the description of infrequent verb valences given by the
training dictionary is less detailed. In particular, the gold-standard
dictionary fails to cover less frequent arguments that are harder to
extract. Unfortunately, the small size of our training and test data
does not enable efficient exploration of how thresholds $p_a$ could
depend on the frequency of the verb.  According to Table I, about half
of the argument types were acknowledged in the test data for just
a~few verbs.

The arguments that we found particularly hard to extract are the
adverbs (\znacz{adv}), with inequality $\text{P}> \text{GSP}$, and
a~group of arguments with $\text{P}$ much smaller than
$\text{GSP}$. The latter include several adjunct-like prepositional
phrases (e.g., \znacz{w+np(loc)}, \lemat{w} means \lemat{in}), certain
clauses (\znacz{PZ} and \znacz{BY}), and the possible lack of subject
\znacz{np(nom)} (=~non-required \znacz{np(nom)}), which corresponds
roughly to the English expletive \lemat{it}.  The inequality
$\text{P}> \text{GSP}$ for adverbs probably reflects their
inconsistent recognition as verb arguments in the gold standard.

The climbing of clitics and objects was another important problem that
we came across when we studied concrete false positives. Namely, some
arguments of the Polish infinitive phrase required by a~finite verb
can be placed anywhere in the sentence. In contrast to Romance
languages, this phenomenon concerns not only clitics. Unfortunately,
\citeauthor{Swidzinski92}'s grammar does not model either object or
clitic climbing and this could have caused the following FPs:
\begin{itemize}
\item 4 of 9 outcomes for \znacz{np(acc)}:
\lemat{kaza\'c} (=~\lemat{order}),
\lemat{m\'oc} (=~\lemat{may}),
\lemat{musie\'c} (=~\lemat{must}),
\lemat{stara\'c (si\k{e})} (=~\lemat{make efforts}),
\item 3 of 11 outcomes for \znacz{np(dat)}:
\lemat{m\'oc},
\lemat{pragn\k{a}\'c} (=~\lemat{desire/wish}),
\lemat{stara\'c (si\k{e})}.
\end{itemize}
There were no FPs that could be attributed to the climbing of the
reflexive marker \lemat{si\k{e}}, although this clitic climbs most
often. For no clear reason, the optimal threshold $p_{a}$ for
\lemat{si\k{e}} was much higher for the training dictionary than for
the test dictionary.

These three frequent arguments also featured relatively many FPs that
were due to omissions in the test dictionary:
\begin{itemize}
\item 1 of 9 outcomes for \znacz{np(acc)}:
\lemat{skar\.zy\'c}  (=~\lemat{accuse}),
\item all outcomes for \lemat{si\k{e}}:
\lemat{pogorszy\'c} (=~\lemat{make worse}),
\lemat{przyzwyczaja\'c} (=~\lemat{get used}),
\lemat{wylewa\'c} (=~\lemat{pour out}),
\lemat{zwi\k{a}za\'c} (=~\lemat{bind}),
\item 6 of 11 outcomes for \znacz{np(dat)}:
\lemat{ciec} (=~\lemat{flow}),
\lemat{dostosowa\'c} (=~\lemat{adjust}),
\lemat{dr\.ze\'c} (=~\lemat{thrill}),
\lemat{d\'zwiga\'c} (=~\lemat{carry}),
\lemat{ratowa\'c} (=~\lemat{save}),
\lemat{wsadzi\'c} (=~\lemat{put into}).
\end{itemize}
As we can see, almost all FPs for these arguments are connected either
to clitic and object climbing or to omissions in the test set.  There
is room for substantial improvement both in the initial corpus parsing
and in the test dictionaries.

\subsection{Evaluation of the co-occurrence matrix adjustment}
\label{ssecEvalCoo}

We obtained the following agreement scores for the three methods of
co-occurrence matrix adjustment defined in Section \ref{subsecMatrix}:
\begin{center}
\begin{tabular}{l|c}
  & agreement score\\
  \hline
  method (A) --- no adjustment (baseline) & $77\%$ \\
  method (B) --- verb-independent matrices & $80\%$ \\
  method (C) --- a~combination of those & $83\%$ 
\end{tabular}  
\end{center}
The scores are statistics (\ref{AgrRate}) computed on the 201 test
verbs for the dictionary of \citet{Swidzinski94} and the preliminary
dictionary processed until Step 6.  Method (C) gave the best results
so it is the only method considered subsequently.

In more detail, Table \ref{tabEval} presents scores for all manually
compiled dictionaries and the automatically extracted dictionary at
several stages of filtering: AE is the preliminary dictionary, AE-A is
the dictionary after correcting the argument sets (Step 3), AE-C is
the one where co-occurrence matrices were corrected using method (C)
(Step 6), and AE-F is the baseline filtered only with the frame-based
binomial hypothesis test (\ref{FraThreshold}). We have constructed
several dictionaries derived from these, such as set-theoretic unions,
intersections, or majority voting, but present only the best
result---the AE-C$+$F, which is the union of frames from the two-stage
filtered AE-C and the one-stage filtered AE-F. The displayed MV is the
majority voting of \citeauthor{Banko00}, \citeauthor{Polanski80}, and
\citeauthor{Swidzinski94}, which are denoted as Ba\'n., Pol., and
\'Swi.

Each cell of two triangular sections of Table \ref{tabEval} presents
the number of pairs, (verb,~frame) or (verb,~argument), that appear
simultaneously in two dictionaries specified by the row and column
titles counted for the 201 test verbs. The displayed recall,
precision, and $F$-score were computed against the MV dictionary.
Recall and precision against other dictionaries can be computed from
the numbers given in the triangular sections.

Although a~large variation of precision and recall can be observed in
Table \ref{tabEval}, the $F$-scores do not vary so much. Assuming the
$F$-score as an objective to be maximized, the two-stage filtering is
better than the frame-based BHT. Namely, we have $F=42\%$ for the AE-C
whereas $F=39\%$ for the AE-F, the scores referring to pairs
(verb,~frame). The set-theoretic union of both dictionaries, AE-C$+$F,
exhibits even a~larger $F=45\%$. In the case of not displayed
dictionaries, we have observed the following triples of
recall/precision/$F$-score: (a) $20\%$/$81\%$/$32\%$ for the
intersection of AE-A, AE-C, and AE-F, (b) $33\%$/$61\%$/$43\%$ for
their majority voting, (c) $39\%$/$45\%$/$42\%$ for their union, and
(d) $39\%$/$46\%$/$42\%$ for the union of just AE-A and AE-F.

The precision of both AE-C and AE-F with respect to the MV is equal to
or higher than that of manually edited dictionaries, whether we look
at single arguments or at frames. A~word of caution is in order,
however.  Very high precision against the MV test dictionary, provided
the recall is sufficient, is a~desirable feature of the automatically
extracted dictionary.  The converse should be expected for the
contributing sources of the MV dictionary. These should be favoured for
presenting frames not occurring in other sources provided all frames
are true. Formally, the contributing sources should feature
very high recall and relatively lower precision against their MV
aggregate.  Exactly this can be observed in Table \ref{tabEval}.

%

\begin{table*}
\begin{tabular}{l|ccccc|cccc}
(verb,~frame)	&AE	&AE-A	&AE-C	&AE-C+F	&AE-F	&Ba\'n.	&Pol.	&\'Swi.	&MV\\
\hline
AE	&7877	&	&	&	&	&	&	&	&\\
AE-A	&848	&983	&	&	&	&	&	&	&\\
AE-C	&587	&645	&658	&	&	&	&	&	&\\
AE-C+F	&675	&674	&658	&746	&	&	&	&	&\\
AE-F	&413	&354	&325	&413	&413	&	&	&	&\\
Ba\'n.	&857	&494	&418	&469	&311	&1660	&	&	&\\
Pol.	&699	&415	&359	&400	&275	&778	&1536	&	&\\
\'Swi.	&697	&409	&363	&406	&294	&766	&778	&1374	&\\
MV	&701	&444	&394	&441	&311	&992	&1004	&992	&1218\\
\hline
recall	&0.58	&0.36	&0.32	&0.36	&0.26	&0.81	&0.82	&0.81	&\\
precision	&0.09	&0.45	&0.60	&0.59	&0.75	&0.6	&0.65	&0.72	&\\
F	&0.16	&0.40	&0.42	&0.45	&0.39	&0.69	&0.73	&0.76	&\\
\end{tabular}

\bigskip
\begin{tabular}{l|ccccc|cccc}
(verb,~argument)	&AE	&AE-A	&AE-C	&AE-C+F	&AE-F	&Ba\'n.	&Pol.	&\'Swi.	&MV\\
\hline
AE	&4051	&	&	&	&	&	&	&	&\\
AE-A	&687	&687	&	&	&	&	&	&	&\\
AE-C	&674	&674	&674	&	&	&	&	&	&\\
AE-C+F	&735	&680	&674	&735	&	&	&	&	&\\
AE-F	&582	&527	&521	&582	&582	&	&	&	&\\
Ba\'n.	&1093	&611	&603	&639	&524	&1342	&	&	&\\
Pol.	&1033	&593	&586	&623	&520	&966	&1336	&	&\\
\'Swi.	&988	&589	&581	&618	&521	&907	&963	&1265	&\\
MV	&1007	&608	&600	&638	&530	&1066	&1122	&1063	&1222\\
\hline
recall	&0.82	&0.50	&0.49	&0.52	&0.43	&0.87	&0.92	&0.87	&\\
precision	&0.25	&0.89	&0.89	&0.87	&0.91	&0.79	&0.84	&0.84	&\\
F	&0.38	&0.64	&0.63	&0.65	&0.58	&0.83	&0.88	&0.85	&\\
\end{tabular}

\caption{
  The comparison of all dictionaries.
}
  \label{tabEval}
\end{table*}

In general, through the correction of co-occurrence matrices in Step 5
and the frame reconstruction (\ref{OArg}), more frames are deleted from
the AE-A dictionary than added. The AE-A contains 338 pairs
(verb,~frame) which do not appear in the obtained AE-C dictionary,
whereas only 13 such pairs from the AE-C are missing in the AE-A. The
sets of pairs (verb,~argument) are almost the same for both
dictionaries.

A~problem that is buried in the apparently good-looking statistics is
the actual shape of co-occurrence matrices in the AE-C dictionary.  In
Step 5 of dictionary filtering, many matrix cells are reset as
independent of the verb. This affects verbs such as \lemat{dziwi\'c}
(=~\lemat{surprise/wonder}). The correct set of frames for this verb
is close to
\begin{align}
  \label{Surprise}
 \FRA(\lemat{dziwi\'c})=
  \klam{
    \begin{array}[c]{l}
    \klam{\znacz{np(nom)},\znacz{np(acc)}},\\
    \klam{\znacz{ZE},\znacz{np(acc)}},\\
    \klam{\znacz{np(nom)},\znacz{sie}},\\
    \klam{\znacz{np(nom)},\znacz{sie},\znacz{np(dat)}},\\
    \klam{\znacz{np(nom)},\znacz{sie},\znacz{ZE}}\\
    \end{array}
  } 
  .
\end{align}
The subordinate clause \znacz{ZE} excludes subject \znacz{np(nom)}
when \lemat{si\k{e}} is missing but it excludes direct object
\znacz{np(acc)} when \lemat{si\k{e}} is present (for there is
a~reflexive diathesis, \lemat{dziwi\'c si\k{e}}=\lemat{be surprised}).

The reconstruction (\ref{OArg}) does not recover the frame set
(\ref{Surprise}) properly for two reasons. Firstly, clause \znacz{ZE}
excludes \znacz{np(acc)} and implies \znacz{np(nom)} for the majority
of verbs.  Secondly, the co-occurrence matrix formalism cannot model
any pairwise exclusion that is conditioned on the absence or presence
of another argument.  However, we suppose that such an argument
interaction is very rare and this deficiency is not so important en
masse.

\subsection{Comparison with previous research}

The scores reported in the literature of verb valence extraction are
so varied that fast conclusions should not be drawn from just a single
figure. For example, \citet{Brent93} achieved $60\%$ recall and $96\%$
precision in the unsupervised approach.  This was done for English and
for a~very small set of extracted valence frames (the set counted only
6 distinct frames).
English-based researchers that evaluated their extracted valence
dictionaries against more complex test dictionaries reported the
following pairs of recall/precision: $36\%/66\%$
\citep{BriscoeCarroll97} against the COMLEX and ANLT dictionaries,
$43\%/90\%$ \citep{Manning93} against \emph{The Oxford Advanced
  Learner's Dictionary}, and $75\%/79\%$ \citep{CarrollRooth98}
against the same dictionary.  

Other factors matter as well. \citeauthor{Korhonen02}
(\citeyear{Korhonen02}, page 77) demonstrated that the results depend
strongly on the filtering method: BHT gives $56\%/50\%$, LLR ---
$48\%/42\%$, MLE --- $58\%/75\%$, no filtering --- $84\%/24\%$, all
methods being frame-based and applied to the same English data.  For
Czech, a~close relative of Polish, \citet{SarkarZeman00} found the
recall/precision pair $74\%/88\%$ but these were evaluated against
a~manually annotated sample of texts rather than against
a~gold-standard valence dictionary. Moreover,
\citeauthor{SarkarZeman00} acquired valence frames from a~manually
disambiguated treebank rather than from raw data, so automatic parsing
did not contribute to the overall error rate.

The closest work to ours is \citet{PrzepiorkowskiFast05b}, who
regarded their own work as preliminary.  They also processed only
a~small part of the 250-million-word IPI PAN Corpus.  Approximately 12
million running words were parsed but sentence parsing was done with
a~simple 18-rule regular grammar rather than with
\citeauthor{Swidzinski94}'s grammar.  Moreover, the dictionary
filtering was done according to several frame-based methods discussed
in the literature and the reference dictionary used was only a~small
part of \citet{Swidzinski94}---100 verbs for a~training set and
another 100 verbs for a~test set.  In contrast to our experiment,
\citeauthor{PrzepiorkowskiFast05b} extracted only non-subject NPs and
PPs.  They ignored subjects, \znacz{np(nom)}, since almost all verbs
subcategorize for them.  The best score in the complete frame
extraction they reported was $48\%$ recall and $49\%$ precision
($F=48\%$), which was obtained for the supervised version of the
binomial hypothesis test (\ref{FraThreshold}).

So as to come closer to the experimental setup of
\citeauthor{PrzepiorkowskiFast05b}, we reapplied all frame filtering
schemes to the case when only non-subject NPs and PPs were retained in
the preliminary dictionary AE and the three manually edited
dictionaries.  The statistics are provided in Table
\ref{tabEvalNPPP}. Under these conditions our two-stage filtering
method added to the frame-based BHT is better again than any of these
methods separately; $F=57\%$ for the AE-C$+$F vs.\ $F=53\%$ for both
the AE-F and AE-C. The AE-C$+$F is not only better than the AE-F and
AE-C with respect to $F$-score but it also contains $15\%$ to $38\%$
more frames. Much higher precision of all these dictionaries than
reported by \citet{PrzepiorkowskiFast05b} may be attributed to the
deep sentence parsing with \'Swigra and the EM disambiguation. The
best recall remains almost the same ($47\%$) for the AE-C$+$F
dictionary, although we extracted valences from a~four fold smaller
amount of text.

\begin{table*}
\begin{tabular}{l|ccccc|cccc}
(verb,~frame)	&AE	&AE-A	&AE-C	&AE-C+F	&AE-F	&Ba\'n.	&Pol.	&\'Swi.	&MV\\
\hline
AE	&3746	&	&	&	&	&	&	&	&\\
AE-A	&695	&713	&	&	&	&	&	&	&\\
AE-C	&533	&539	&544	&	&	&	&	&	&\\
AE-C+F	&615	&585	&544	&626	&	&	&	&	&\\
AE-F	&453	&417	&371	&453	&453	&	&	&	&\\
Ba\'n.	&827	&481	&407	&463	&377	&1255	&	&	&\\
Pol.	&693	&426	&367	&412	&338	&684	&1128	&	&\\
\'Swi.	&645	&422	&368	&413	&346	&662	&661	&939	&\\
MV	&694	&455	&395	&446	&372	&820	&819	&797	&955\\
\hline
recall	&0.73	&0.48	&0.41	&0.47	&0.39	&0.86	&0.86	&0.83	&\\
precision	&0.19	&0.64	&0.73	&0.71	&0.82	&0.65	&0.73	&0.85	&\\
F	&0.30	&0.55	&0.53	&0.57	&0.53	&0.74	&0.79	&0.84	&\\
\end{tabular}

\bigskip
\begin{tabular}{l|ccccc|cccc}
(verb,~argument)	&AE	&AE-A	&AE-C	&AE-C+F	&AE-F	&Ba\'n.	&Pol.	&\'Swi.	&MV\\
\hline
AE	&2364	&	&	&	&	&	&	&	&\\
AE-A	&392	&392	&	&	&	&	&	&	&\\
AE-C	&385	&385	&385	&	&	&	&	&	&\\
AE-C+F	&415	&388	&385	&415	&	&	&	&	&\\
AE-F	&354	&327	&324	&354	&354	&	&	&	&\\
Ba\'n.	&717	&353	&349	&369	&322	&881	&	&	&\\
Pol.	&659	&333	&330	&346	&306	&603	&813	&	&\\
\'Swi.	&585	&323	&319	&334	&296	&547	&567	&715	&\\
MV	&633	&346	&342	&360	&317	&665	&685	&629	&747\\
\hline
recall	&0.85	&0.46	&0.46	&0.48	&0.42	&0.89	&0.92	&0.84	&\\
precision	&0.27	&0.88	&0.89	&0.87	&0.90	&0.75	&0.84	&0.88	&\\
F	&0.41	&0.60	&0.61	&0.62	&0.57	&0.81	&0.88	&0.86	&\\
\end{tabular}

\caption{
  The case of source dictionaries restricted to non-subject NPs and PPs.
}
  \label{tabEvalNPPP}
\end{table*}

\section{Conclusion}
\label{secConclusion}

Two new ideas for valence extraction have been proposed and applied to
Polish language data in this paper.  Firstly, we have introduced
a~two-step scheme for filtering incorrect frames. The list of valid
arguments was determined for each verb first and then a~method of
combining arguments into frames was found. The two-stage induction was
motivated by an observation that the argument combination rules, such
as co-occurrence matrices, are largely independent of the verb. We
suppose that this observation is not language-specific and the
co-occurrence matrix formalism can be easily tailored to improve verb
valence extraction for many other languages and special datasets (also
subdomain corpora and subdomain valence dictionaries).  The second new
idea is a~simple EM selection algorithm, which is a~natural baseline
method for unsupervised disambiguation tasks such as choosing the
correct valence frame for a~sentence. In our application it helped
high-precision valence extraction without a~large treebank or
a~probabilistic parser.

Although the proposed frame filtering technique needs further work to
address the drawbacks noticed in Subsection \ref{ssecEvalCoo} and to
improve the overall performance, the present results are encouraging
and suggest that two-step frame filtering is worth developing. In
future work, experiments can be conducted using various schemes of
decomposing the information contained in the sets of valence frames
and, due to the scale of the task, this decomposition should be done
to a~large extent in an algorithmic way.  The straightforward idea to
explore is to express the verb valence information in terms of $n$-ary
rather than binary relations among verbs and verb arguments, where
$n>2$. Subsequently, one can investigate the analogous learning
problem and propose a~frame-set reconstruction scheme for the $n$-ary
relations.  Are ternary relations sufficient to describe the valence
frame sets?  We disbelieve that relations of irreducibly large arities
appear in human language lexicons since, for example,
\citet{HalfordWilsonPhillips98} observed that human capacity for
processing random $n$-ary relations depends strongly on the relation
arity.

Knowing algebraic constraints on the verb argument combinations is
important also for language resource maintenance.  Because our test
dictionaries do not list valid argument combinations extensively, many
false positive frames in the two-stage corrected dictionary were in
fact truly positive. Thus, it is advisable to correct gold-standard
dictionaries themselves, for example using a~modification of the
reconstruction (\ref{OArg}). However, prior to resetting the
gold-standard in this way, it must be certain that the reconstruction
process does not introduce linguistically implausible frames. Also for
this reason, the effective complexity of verb-argument and
argument-argument relations in natural language should be investigated
thoroughly from a~more mathematical point of view.

\appendix

\section{A faster reconstruction of the frame set}
\label{appMatrix}

Although there is no need to compute $\bar\FRA(v)$ defined in
(\ref{OArg}) to verify condition $f\in\bar\FRA(v)$ for a~given $f$,
the reconstruction $\bar\FRA(v)$ can be computed efficiently if needed
for other purposes. A~naive solution suggested by formula (\ref{OArg})
is to search through all elements of the power set $2^{\LIS(v)}$ and
to check for each independently whether it is an element of
$\bar\FRA(v)$.  However, we can do it faster by applying some dynamic
programming.

Firstly, let us enumerate the elements of
$\LIS(v)=\klam{b_1,b_2,...,b_N}$. In the following, we will compute
the chain of sets $A_0,A_1,...,A_N$ where
  $A_n=\klam{
    (B_n\cap f,
    B_n\setminus f)
    \middle| f\in \bar\FRA(v)
  }$
and $B_n =\klam{b_1,b_2,...,b_n}$.

In fact, there is an iteration for this chain:
\begin{align*}
 A_0&=\klam{(\emptyset,\emptyset)},
 \\
 A_n&=
 \klam{(f\cup \klam{b_n}, g) \middle| 
    \begin{array}[c]{l}
   (f,g)\in A_{n-1}
   ,\\
   \forall_{a\in f}\, \MAT(v)_{b_na}\not=\excl
   ,\\
   \forall_{a\in g}\, \MAT(v)_{b_na}\not=\earr
   ,\\
   \forall_{a\in g}\, \MAT(v)_{b_na}\not=\larr
    \end{array}
 }
 \\
 &\phantom{==} 
 \cup \klam{(f, g\cup \klam{b_n}) \middle| 
    \begin{array}[c]{l}
   (f,g)\in A_{n-1}
   ,\\
   \klam{b_n}\not\in \EXP(v)
   ,\\
   \forall_{a\in f}\, \MAT(v)_{b_na}\not=\earr
   ,\\
   \forall_{a\in f}\, \MAT(v)_{b_na}\not=\larr
    \end{array}
 }
 .
\end{align*}
Once the set $A_N=\klam{ (f, \LIS(v)\setminus f) \middle| f\in \bar\FRA(v) }$
is computed, $\bar\FRA(v)$ can be read off easily.

\section{Parsing of the IPI PAN Corpus}
\label{appParsing}

The input of the valence extraction experiment discussed in this paper
came from the 250-million-word IPI PAN Corpus of Polish
(\url{http://korpus.pl/}).  The original automatic part-of-speech
annotation of the text was removed, since it contained too many
errors, and the sentences from the corpus were analyzed using the
\'Swigra parser (\citealp{Wolinski04}, \citeyear{Wolinski05}), see
also \url{http://nlp.ipipan.waw.pl/~wolinski/swigra/}. Technically,
\'Swigra utilizes two distinct language resources: (1)
Morfeusz---a~dictionary of inflected words (a.k.a.\ a morphological
analyzer) programmed by \citet{Wolinski06} on the basis of about
20,000 stemming rules compiled by \citet{Tokarski93}, and (2)
GFJP---the formal grammar of Polish written by \citet{Swidzinski92}.
\citeauthor{Swidzinski92}'s grammar is a~DCG-like grammar, close to
the format of the metamorphosis grammar by \citet{Colmerauer78}.  It
counts 461 rules and examples of its parse trees can be found in
\citet{Wolinski04}. For the sake of this project, \'Swigra used a~fake
valence dictionary that allowed any verb to take none or one NP in the
nominative (the subject) and any combination of other arguments.

Only a~small subset of sentences was actually selected to be parsed
with \'Swigra.  The following selection criteria were applied to the
whole 250-million-word IPI PAN Corpus:
\begin{enumerate}
\item The selected sentence had to contain a~word recognized by
  Morfeusz as a~verb and the verb had to occur $\ge 396$ times in the
  corpus. ($396$ is the lowest corpus frequency of a~verb from the
  test set described in Section \ref{secEvaluation}. The threshold was
  introduced to speed up parsing without loss of empirical coverage
  for any verb in the test set. The selected sentence might contain
  another less frequent verb if it was a~compound sentence.)
\item The selected sentence could not be longer than 15 words.  (We
  supposed that the EM selection would find it difficult to select the
  correct parse for longer sentences.)
\item Maximally 5000 sentences were selected per recognized verb.  (We
  supposed that a~frame which was used less than once per one 5000
  verb occurrences would not be considered in the gold-standard
  dictionaries.)
\end{enumerate}
In this way, a~subset of $1\,011\,991$ sentences ($8\,727\,441$
running words) was chosen. They were all fed to \'Swigra's input but
less than half (0.48 million sentences) were parsed successfully
within a preset time of 1 minute per sentence.  Detailed statistics
are given in Table \ref{tabCorpusSizes} below. All mentioned
thresholds were introduced in advance to compute only the most useful
parse forests in the pre-estimated total time of a~few months. It was
the first experiment ever in which \'Swigra was applied to more than
several hundred sentences.  The parsing actually took 2 months on
a~single PC station.

Not all information contained in the obtained parse forests was
relevant for valence acquisition.  Full parses were subsequently
reduced to valence frames plus verbs, as in the first displayed
example in Section \ref{secExtraction}. First of all, the parse
forests for compound sentences were split into separate parse forests
for elementary clauses.  Then each parse tree was reduced to a~string
that identifies only the top-most phrases.  To decrease the amount of
noise in the subsequent EM selection and to speed up computation, we
decided to skip 10\% of clauses that had the largest number of
reduced parses. As a~result, we only retained clauses which had $\le
40$ reduced parses.

To improve the EM selection, we also deleted parses that contained
certain syntactically idiosyncratic words---mostly indefinite pronouns
\lemat{to} (=~\lemat{this}), \lemat{co} (=~\lemat{what}), and
\lemat{nic} (=~\lemat{nothing})---or highly improbable morphological
word interpretations (like the second interpretation for
\lemat{albo}~=~1.~the conjunction \lemat{or}; 2.  the vocative
singular of the noun \lemat{alb}---a kind of liturgical vestment). The
stop list of improbable interpretations consisted of 646 word
interpretations which never occurred in the SFPW Corpus but were
possible interpretations of the most common words according to
Morfeusz. The SFPW Corpus is a~manually POS tagged 0.5-million-word
corpus prepared for the frequency dictionary of 1960s Polish
\citep{KurczOther90}, which was actually commenced in the 1960s but
not published until 1990.

Our format of reduced parses approximates the format of valence frames
in \citet{Swidzinski94}, so it diverges from the format proposed by
\citet{Przepiorkowski06}. To convert a~parse in Przepiorkowski's
format into ours, the transformations must be performed as follows:
\begin{enumerate}
\item Add the dropped personal subject or the impersonal subject
  expressed by the ambiguous reflexive marker \lemat{si\k{e}} when
  their presence is implied by the verb form.
\item Remove one nominal phrase in the genitive for negated verbs. (An
  attempt to treat the genitive of negation.)
\item Transform several frequent adjuncts expressed by
  nominal phrases.
\item Skip the parse if it contains pronouns \lemat{to}
  (=~\lemat{this}), \lemat{co} (=~\lemat{what}), and \lemat{nic}
  (=~\lemat{nothing}). (Instead of converting these pronouns
  into regular nominal phrases.)
\item Remove lemmas from non-verbal phrases and sort phrases in
  alphabetic order.
\end{enumerate}

The resulting bank of reduced parse forests included $510\,743$
clauses with one or more proposed valence frames. We parsed
successfully only 3.4 million running words of the whole
250-million-word IPI PAN Corpus---four times less than the 12 million
words parsed by \citet{PrzepiorkowskiFast05b}. However, our superior
results in the valence extraction task indicate that skipping
a~fraction of available empirical data is a~good idea if the remaining
data can be processed more thoroughly and the skipped portion does not
provide different efficiently usable information.

\begin{table*}
\begin{tabular}{l|rr}
& sentences/clauses & words \\
\hline 
sentences sent to \'Swigra's input 
& 1\,011\,991 sentences & 8\,727\,441 \\
sentences successfully parsed with \'Swigra 
& 481\,039 sentences & 3\,421\,863 \\
sentences with $\le 40$ parses split into clauses 
& 569\,307  clauses & 3\,149\,391 \\
the final bank of reduced parse forests 
& 510\,743  clauses & 2\,795\,357 \\
\end{tabular}

\caption{
  Sizes of the processed parts of the IPI PAN Corpus.
}
  \label{tabCorpusSizes}
\end{table*}

\section{The EM selection algorithm}
\label{appEM}

Consider the following abstract statistical task.  Let
$Z_1,Z_2,...,Z_M$, with $Z_i:\Omega\rightarrow J$, be a~sequence of
discrete random variables and let $Y_1,Y_2,...,Y_M$ be a~random sample
of sets, where each set $Y_i:\Omega\rightarrow 2^J\setminus\emptyset$
contains the actual value of $Z_i$, i.e., $Z_i\in Y_i$.  The objective
is to guess the conditional distribution of $Z_i$ given an event
$(Y_i=A_i)_{i=1}^M$, $A_i\subset J$. In particular, we would like to
know the conditionally most likely values of $Z_i$.  The exact
distribution of $Y_i$ is not known and unfeasible to estimate if we
treat the values of $Y_i$ as atomic entities. We have to solve the
task via some rationally motivated assumptions.

Our heuristic solution was iteration
\begin{align}
\label{EM1a}
  p_{ji}^{(n)}&=
  \begin{cases}
    p_j^{(n)}/\sum_{j'\in A_i} p_{j'}^{(n)}, & j\in A_i,
    \\
    0, & \text{else},    
  \end{cases}
  \\
  \label{EM2a}
  p_j^{(n+1)}&= \frac{1}{M} \sum_{i=1}^{M} p_{ji}^{(n)}
  ,
\end{align}
with $p_j^{(1)}=1$. We observed that coefficients $p_{ji}^{(n)}$
converge to a~value that can be plausibly identified with the
conditional probability $P(Z_i=j|Y_i=A_i)$.  

Possible applications of iteration (\ref{EM1a})--(\ref{EM2a}), which
we call the EM selection algorithm, cover unsupervised disambiguation
tasks where the number of different values of $Y_i$ is very large but
the internal ambiguity rate (i.e., the typical cardinality
$\card{Y_i}$) is rather small and the alternative choices within $Y_i$
(i.e., the values of $Z_i$) are highly repeatable. There may be many
applications of this kind in NLP and bioinformatics. To our knowledge,
however, we present the first rigorous treatment of this particular
selection problem.

In this appendix, we will show that the EM selection algorithm belongs
to the class of expectation-maximization (EM) algorithms.  For this
reason, our algorithm resembles many instances of EM used in NLP, such
as the Baum-Welch algorithm for hidden Markov models
\citep{Baum72} or linear interpolation \citep{Jelinek97}. However,
normalization (\ref{EM1a}), which is done over varying sets
$A_i$---unlike the typical case of linear interpolation, is the
singular feature of EM selection. The local maxima of the respective
likelihood function also form a~convex set, so there is no need to
care much for initializing the iteration (\ref{EM1a})--(\ref{EM2a}),
unlike e.g.\ the Baum-Welch algorithm.

To begin with, we recall the universal scheme of EM
\citep{DempsterLairdRubin77,NealHinton99}.  Let $P(Y|\theta)$ be
a~likelihood function, where $Y$ is an observed variable and
$\theta$ is an unknown parameter.
For the observed value $Y$, the maximum
likelihood estimator of $\theta$ is 
\begin{align*}
  \theta_{\text{MLE}}=\argmax_\theta P(Y|\theta).
\end{align*}
When the direct maximization is impossible, we may consider a~latent
discrete variable $Z$ and function 
\begin{align*}
  Q(\theta',\theta'')
  &=\sum_{z} P(Z=z|Y,\theta')\log P(Z=z,Y|\theta''),
\end{align*}
which is a~kind of cross entropy function.  The EM algorithm consists
of setting an initial parameter value $\theta_1$ and iterating
\begin{align}
\label{EM}
  \theta_{n+1}=\argmax_\theta Q\okra{\theta_{n},\theta}
\end{align}
until a~sufficient convergence of $\theta_{n}$ is achieved. It is
a~general fact that $P\okra{Y|\theta_{n+1}}\ge P\okra{Y|\theta_{n}}$
but EM is worth considering only if maximization (\ref{EM}) is easy.

Having outlined the general EM algorithm, we come back to the
selection problem. The observed variable is $Y=(Y_1,Y_2,...,Y_M)$, the
latent one is $Z=(Z_1,Z_2,...,Z_M)$, whereas the parameter seems to be
$\theta_n=\okra{p_j^{(n)}}_{j\in J}$. The appropriate likelihood
function remains to be determined.  We may suppose from the problem
statement that it factorizes into $P(Z,Y|\theta) =\prod_{i}
P(Z_i,Y_i|\theta)$.  Hence $Q(\theta',\theta'')$ takes the form
\begin{align*}
  Q(\theta',\theta'')
  &=\sum_{i} \sum_{j} P(Z_i=j|Y_i=A_i,\theta')  
  \log P(Z_i=j,Y_i=A_i|\theta''). 
\end{align*}

Assume now
\begin{align}
  \label{LikeYonZ}
   P(Y_i=A|Z_i=j,\theta)&=
  \begin{cases}
    g(A), & j\in A,
    \\
    0, & \text{else},
  \end{cases}
  \\
  \label{LikeZ}
  P(Z_i=j|\theta)&=p_j
\end{align}
for $\theta=\okra{p_j}_{j\in J}$ and a~parameter-free function $g(\cdot)$
satisfying
\begin{align}
\label{gConstraint}
   \sum_{A\in 2^J} \boole{j\in A} g(A) =1,  \quad\forall {j\in J},
\end{align} 
where
$$
\boole{\phi}=
\begin{cases}
1, & \text{$\phi$ is true},
\\
0, & \text{else}.
\end{cases}
$$
For example, let $g(A)=q^{\card{A}-1} (1-q)^{\card{J}-\card{A}}$,
where $\card{A}$ stands for the cardinality of set $A$ and $0\le q\le
1$ is a~fixed number not incorporated into $\theta$. Then the
cardinalities of sets $Y_i$ are binomially distributed, i.e.,
$P(\card{Y_i}-1|\theta)\sim B(\card{J}-1,q)$. This particular form of
$g(A)$, however, is not necessary to satisfy (\ref{gConstraint}).

The model (\ref{LikeYonZ})--(\ref{LikeZ}) is quite speculative. In the
main part of this article, we need to model the probability
distribution of the reduced parse forest $Y_i$ under the assumption
that the correct parse $Z_i$ is an arbitrary element of $Y_i$. In
particular, we have to imagine what $P(Y_i=A|Z_i=j,\theta)$ is like if
$j$ is a~semantically implausible parse. We circumvent the difficulty
by saying in (\ref{LikeYonZ}) that this quantity is the same as if $j$
were the correct parse.

Assumption (\ref{LikeYonZ}) leads to an EM algorithm which does not
depend on the specific choice of function $g(\cdot)$.  Therefore the
algorithm is rather generic. In fact, (\ref{LikeYonZ}) assures that
$P(Y_i=A_i|\theta)=g(A_i)P(Z_i\in A_i|\theta)$ and
\begin{align}
  \label{ConditionalZ}
  P(Z_i=j|Y_i=A_i,\theta)&=P(Z_i=j|Z_i\in A_i,\theta).
\end{align}
In consequence, iteration (\ref{EM}) is equivalent to
\begin{align}
  0&= \subst{
    \frac{\partial}{\partial p_j}
    \kwad{Q(\theta_n,\theta)-\lambda \okra{\sum_{j'\in J} p_{j'} -1}}
  }_{\theta=\theta_{n+1}}
  = \frac{\sum_{i=1}^{M}  p_{ji}^{(n)}}{p_j^{(n+1)}}-\lambda
  ,
  \label{LagrangeN}
\end{align}
where $p_{ji}^{(n)}=P(Z_i=j|Z_i\in A_i,\theta_{n})$ is given
exactly by (\ref{EM1a}). 

If the Lagrange multiplier $\lambda$ is assigned the value that satisfies
constraint $\textstyle\sum_{j\in J} p_{j'}=1$ then equation
(\ref{LagrangeN}) simplifies to (\ref{EM2a}).  Hence it becomes
straightforward that iteration (\ref{EM1a})--(\ref{EM2a}) maximizes
locally the log-likelihood
\begin{align}
  \label{PYN}
  L(\theta):=\log P((Y_i=A_i)_{i=1}^M|\theta)&
  =\log\kwad{\prod_{i=1}^M \frac{P(Z_i\in A_i|\theta)}{g(A_i)}}
  ,
\end{align}
or simply $L^{(n+1)}\ge L^{(n)}$ for 
\begin{align*}
  L^{(n)}:=L(\theta_n) + \sum_{i=1}^{M} \log g(A_i)
  =\sum_{i=1}^{M} \log \kwad{\sum_{j\in A_i} p_j^{(n)}}  
  ,
  \quad
  n\ge 2
  .
\end{align*}

Moreover, there is no need to care for the initialization of iteration
(\ref{EM1a})--(\ref{EM2a}) since the local maxima of function
(\ref{PYN}) form a~convex set $\mathcal{M}$, i.e., $\theta,\theta'\in
\mathcal{M}\implies q\theta+(1-q)\theta'\in \mathcal{M}$ for $0\le
q\le 1$. Hence that function is, of course, constant on
$\mathcal{M}$. To show this, observe that the domain of log-likelihood
(\ref{PYN}) is a~convex compact set $\mathcal{P}=\klam{\theta: \sum_j
  p_j=1,\ p_j\ge 0}$.  The second derivative of $L$ reads
\begin{align*}
  L_{jj'}(\theta):=\frac{\partial^2L(\theta)}{\partial p_{j}\partial p_{j'}}
  =-\sum_{i=1}^{M} \frac{
    \boole{j\in A_i}\boole{j'\in A_i}
  }{
    \okra{\sum_{j''\in A_i} p_{j''}}^2
  }
  .
\end{align*}
Since matrix $\klam{L_{jj'}}$ is negative definite, i.e.,
$\sum_{jj'} a_j L_{jj'}(\theta) a_{j'}
\le 0
$,
function $L$ is concave.
As a~general fact, a~continuous function $L$ achieves its supremum on
a~compact set $\mathcal{P}$ (\citealp{Rudin74}, Theorem 2.10).  If
additionally $L$ is concave and its domain $\mathcal{P}$ is convex
then the local maxima of $L$ form a~convex set $\mathcal{M}\subset
\mathcal{P}$, where $L$ is constant and achieves its supremum
(\citealp{BoydVandenberghe04}, Section 4.2.2).


\acknowledgements 

Grateful acknowledgements are due to Marcin Woli\'nski for his help in
using \'Swigra, to Witold Kiera\'s for retyping samples of the test
dictionaries, and to Marek \'Swidzi\'nski for offering the source file
of his valence dictionary.  The author thanks also Adam
Przepi\'orkowski, Jan Mielniczuk, Laurence Cantrill, and the anonymous
reviewers for many helpful comments concerning the composition of this
article. The work was supported by the Polish State Research Project,
3 T11C 003 28, \textit{Automatyczna ekstrakcja wiedzy lingwistycznej z
  du\.zego korpusu j\k{e}zyka polskiego}.


\theendnotes

\bibliographystyle{klunamed}
\bibliography{0-journals-full,0-publishers-full,ai,mine,books,nlp,neuro,tcs}

\begin{thebibliography}{}

\bibitem[\protect\citeauthoryear{Artstein and Poesio}{2008}]{ArtsteinPoesio08}
Artstein, R. and M. Poesio: 2008, `Inter-coder agreement for computational
  linguistics'.
\newblock {\em Computational Linguistics} {\bf 34}, 555--596.

\bibitem[\protect\citeauthoryear{Baker and
  Ruppenhofer}{2002}]{BakerRuppenhofer02}
Baker, C.~F. and J. Ruppenhofer: 2002, `{FrameNet's} Frames vs. {Levin's} Verb
  Classes'.
\newblock In: {\em Proceedings of the 28th Annual Meeting of the Berkeley
  Linguistics Society}.
\newblock pp. 27--38.

\bibitem[\protect\citeauthoryear{Ba\'nko}{2000}]{Banko00}
Ba\'nko, M. (ed.): 2000, {\em Inny s{\l}ownik j\k{e}zyka polskiego}.
\newblock Warszawa: Wydawnictwo Naukowe PWN.

\bibitem[\protect\citeauthoryear{Baum}{1972}]{Baum72}
Baum, L.~E.: 1972, `Inequality and Associated Maximization Technique In
  Statistical Estimation of Probabilistic Functions of {Markov} processes'.
\newblock {\em Inequalities} {\bf 3}, 1--8.

\bibitem[\protect\citeauthoryear{Bennett
  et~al.}{1954}]{BennettAlpertGoldstein54}
Bennett, E.~M., R. Alpert, and A.~C. Goldstein: 1954, `Communications through
  limited questioning'.
\newblock {\em Public Opinion Quarterly} {\bf 18(3)}, 303--308.

\bibitem[\protect\citeauthoryear{Boyd and
  Vandenberghe}{2004}]{BoydVandenberghe04}
Boyd, S. and L. Vandenberghe: 2004, {\em Convex Optimization}.
\newblock Cambridge: Cambridge University Press.

\bibitem[\protect\citeauthoryear{Brent}{1993}]{Brent93}
Brent, M.~R.: 1993, `From Grammar to Lexicon: Unsupervised Learning of Lexical
  Syntax'.
\newblock {\em Computational Linguistics} {\bf 19}, 243--262.

\bibitem[\protect\citeauthoryear{Briscoe and Carroll}{1997}]{BriscoeCarroll97}
Briscoe, T. and J. Carroll: 1997, `Automatic Extraction of Subcategorization
  from Corpora'.
\newblock In: {\em Proceedings of the 5th ACL Conference on Applied Natural
  Language Processing, Washington, DC}.
\newblock Morgan Kaufmann, pp. 356--363.

\bibitem[\protect\citeauthoryear{Carroll and Rooth}{1998}]{CarrollRooth98}
Carroll, G. and M. Rooth: 1998, `Valence Induction with a Head-Lexicalized
  {PCFG}'.
\newblock In: {\em Arbeitspapiere des Instituts f\"ur Maschinelle
  Sprachverarbeitung, no. 4, vol. 3}.
\newblock pp. 25--54.

\bibitem[\protect\citeauthoryear{Chesley and
  Salmon-Alt}{2006}]{ChesleySalmonAlt06}
Chesley, P. and S. Salmon-Alt: 2006, `Automatic extraction of subcategorization
  frames for {French}'.
\newblock In: {\em Proceedings of the Language Resources and Evaluation
  Conference, LREC 2006, Genua, Italy}.

\bibitem[\protect\citeauthoryear{Chi and Geman}{1998}]{ChiGeman98}
Chi, Z. and S. Geman: 1998, `Estimation of probabilistic context-free
  grammars'.
\newblock {\em Computational Linguistics} {\bf 24}, 299--305.

\bibitem[\protect\citeauthoryear{Colmerauer}{1978}]{Colmerauer78}
Colmerauer, A.: 1978, `Metamorphosis grammar'.
\newblock In: {\em Natural Language Communication with Computers}, Lecture
  Notes in Computer Science 63.
\newblock New York: Springer, pp. 133--189.

\bibitem[\protect\citeauthoryear{Dempster et~al.}{1977}]{DempsterLairdRubin77}
Dempster, A.~P., N.~M. Laird, and D.~B. Rubin: 1977, `Maximum Likelihood from
  Incomplete Data via the {EM} algorithm'.
\newblock {\em Journal of the Royal Statistical Society, series B} {\bf 39},
  185--197.

\bibitem[\protect\citeauthoryear{D\k{e}bowski and
  Woli\'nski}{2007}]{DebowskiWolinski07}
D\k{e}bowski, {\L}. and M. Woli\'nski: 2007, `Argument co-occurrence matrix as
  a description of verb valence'.
\newblock In: Z. Vetulani (ed.): {\em Proceedings of the 3rd Language \&
  Technology Conference, October 5-7, 2007, Pozna\'n, Poland}.
\newblock pp. 260--264.

\bibitem[\protect\citeauthoryear{Ersan and Charniak}{1995}]{ErsanCharniak95}
Ersan, M. and E. Charniak: 1995, `A statistical syntactic disambiguation
  program and what it learns'.
\newblock In: S. Wermter, E. Riloff, and G. Scheler (eds.): {\em Learning for
  Natural Language Processing}.
\newblock New York: Springer, pp. 146--159.

\bibitem[\protect\citeauthoryear{Fast and
  Przepi\'orkowski}{2005}]{PrzepiorkowskiFast05b}
Fast, J. and A. Przepi\'orkowski: 2005, `Automatic Extraction of {Polish} Verb
  Subcategorization: An Evaluation of Common Statistics'.
\newblock In: Z. Vetulani (ed.): {\em Proceedings of the 2nd Language \&
  Technology Conference, Pozna\'n, Poland, April 21--23, 2005}.
\newblock pp. 191--195.

\bibitem[\protect\citeauthoryear{Gorrell}{1999}]{Gorrell99}
Gorrell, G.: 1999, `Acquiring Subcategorisation from Textual Corpora'.
\newblock M. Phil. dissertation, University of Cambridge.

\bibitem[\protect\citeauthoryear{Halford
  et~al.}{1998}]{HalfordWilsonPhillips98}
Halford, G.~S., W.~H. Wilson, and W. Phillips: 1998, `Processing capacity
  defined by relational complexity: {Implications} for comparative,
  developmental and cognitive psychology'.
\newblock {\em Behavioral Brain Sciences} {\bf 21(6)}, 803--864.

\bibitem[\protect\citeauthoryear{Jelinek}{1997}]{Jelinek97}
Jelinek, F.: 1997, {\em Statistical Methods for Speech Recognition}.
\newblock Cambridge, MA: The MIT Press.

\bibitem[\protect\citeauthoryear{Korhonen}{2002}]{Korhonen02}
Korhonen, A.: 2002, `Subcategorization Acquisition'.
\newblock Ph. D. dissertation, University of Cambridge.

\bibitem[\protect\citeauthoryear{Kupiec}{1992}]{Kupiec92}
Kupiec, J.: 1992, `Robust part-of-speech tagging using a hidden {Markov}
  model'.
\newblock {\em Computer Speech and Language} {\bf 6}, 225--242.

\bibitem[\protect\citeauthoryear{Kurcz et~al.}{1990}]{KurczOther90}
Kurcz, I., A. Lewicki, J. Sambor, and J. Woronczak: 1990, {\em {S{\l}ownik
  frekwencyjny polszczyzny wsp\'o{\l}czesnej}}.
\newblock Krak\'ow: Instytut J\k{e}zyka Polskiego PAN.

\bibitem[\protect\citeauthoryear{Lapata and Brew}{2004}]{LapataBrew04}
Lapata, M. and C. Brew: 2004, `Verb Class Disambiguation using Informative
  Priors'.
\newblock {\em Computational Linguistics} {\bf 30}, 45--73.

\bibitem[\protect\citeauthoryear{Levin}{1993}]{Levin93}
Levin, B.: 1993, {\em English Verb Classes and Alternations: A Preliminary
  Investigation}.
\newblock Chicago and London: The University of Chicago Press.

\bibitem[\protect\citeauthoryear{Macleod
  et~al.}{1994}]{MacleodGrishmanMeyers94}
Macleod, C., R. Grishman, and A. Meyers: 1994, `Creating a Common Syntactic
  Dictionary of English'.
\newblock In: {\em SNLR: International Workshop on Sharable Natural Language
  Resources, Nara, August, 1994}.

\bibitem[\protect\citeauthoryear{Manning}{1993}]{Manning93}
Manning, C.: 1993, `Automatic acquisition of a large subcategorization
  dictionary from corpora'.
\newblock In: {\em Proceedings of the 31st Annual Meeting of the ACL, Columbus,
  Ohio}.
\newblock pp. 235--242.

\bibitem[\protect\citeauthoryear{Mayol et~al.}{2005}]{MayolBoledaBadia05}
Mayol, L., G. Boleda, and T. Badia: 2005, `Automatic acquisition of syntactic
  verb classes with basic resources'.
\newblock {\em Language Resources and Evaluation} {\bf 39}, 295--312.

\bibitem[\protect\citeauthoryear{McCarthy}{2001}]{McCarthy01}
McCarthy, D.: 2001, `Lexical Acquisition at the Syntax-Semantics Interface:
  Diathesis Alternations, Subcategorization Frames and Selectional
  Preferences'.
\newblock Ph.D. thesis, University of Sussex.

\bibitem[\protect\citeauthoryear{Merialdo}{1994}]{Merialdo94}
Merialdo, B.: 1994, `Tagging {English} text with a probabilistic model'.
\newblock {\em Computational Linguistics} {\bf 20}, 155--171.

\bibitem[\protect\citeauthoryear{M{\l}ynarczyk}{2004}]{Mlynarczyk04}
M{\l}ynarczyk, A.~K.: 2004, `Aspectual Pairing in Polish'.
\newblock Ph.D. thesis, Universiteit Utrecht.

\bibitem[\protect\citeauthoryear{Neal and Hinton}{1999}]{NealHinton99}
Neal, R. and G. Hinton: 1999, `A view of the {EM} algorithm that justifies
  incremental, sparse, and other variants'.
\newblock In: M.~I. Jordan (ed.): {\em Learning in Graphical Models}.
\newblock Cambridge, MA: The MIT Press, pp. 355--368.

\bibitem[\protect\citeauthoryear{Pola\'nski}{1992}]{Polanski80}
Pola\'nski, K. (ed.): 1980--1992, {\em {S{\l}ownik syntaktyczno-generatywny
  czasownik\'ow polskich}}.
\newblock Wroc{\l}aw: Zak{\l}ad Narodowy im. Ossoli\'nskich / Krak\'ow:
  Instytut J\k{e}zyka Polskiego PAN.

\bibitem[\protect\citeauthoryear{Przepi\'orkowski}{2006}]{Przepiorkowski06}
Przepi\'orkowski, A.: 2006, `What to acquire from corpora in automatic valence
  acquisition'.
\newblock In: V. Koseska-Toszewa and R. Roszko (eds.): {\em Semantyka a
  konfrontacja j\k{e}zykowa (3)}.
\newblock Warszawa: Slawistyczny O\'srodek Wydawniczy PAN.

\bibitem[\protect\citeauthoryear{Przepi\'orkowski and
  Fast}{2005}]{PrzepiorkowskiFast05}
Przepi\'orkowski, A. and J. Fast: 2005, `Baseline Experiments in the Extraction
  of {Polish} Valence Frames'.
\newblock In: M.~A. K{\l}opotek, S.~T. Wierzcho\'n, and K. Trojanowski (eds.):
  {\em Intelligent Information Processing and Web Mining}.
\newblock New York: Springer, pp. 511--520.

\bibitem[\protect\citeauthoryear{Przepi\'orkowski and
  Woli\'nski}{2003}]{PrzepiorkowskiWolinski03}
Przepi\'orkowski, A. and M. Woli\'nski: 2003, `A~Flexemic Tagset for {Polish}'.
\newblock In: {\em Proceedings of Morphological Processing of {Slavic}
  Languages, {EACL}~2003}.
\newblock pp. 33--40.

\bibitem[\protect\citeauthoryear{Rudin}{1974}]{Rudin74}
Rudin, W.: 1974, {\em Real and complex analysis}.
\newblock New York: McGraw-Hill.

\bibitem[\protect\citeauthoryear{Sarkar and Zeman}{2000}]{SarkarZeman00}
Sarkar, A. and D. Zeman: 2000, `Automatic Extraction of Subcategorization
  Frames for {Czech}'.
\newblock In: {\em Proceedings of the 18th International Conference on
  Computational Linguistics, COLING 2000, Saarbr{\"u}cken, Germany}.
\newblock pp. 691--698.

\bibitem[\protect\citeauthoryear{{Schulte im Walde}}{2006}]{SchulteImWalde06}
{Schulte im Walde}, S.: 2006, `Experiments on the Automatic Induction of
  {German} Semantic Verb Classes'.
\newblock {\em Computational Linguistics} {\bf 32}, 159--194.

\bibitem[\protect\citeauthoryear{Surdeanu
  et~al.}{2008}]{SurdeanuMoranteMarquez08}
Surdeanu, M., R. Morante, and L. M\`arquez: 2008, `Analysis of Joint Inference
  Strategies for the Semantic Role Labeling of Spanish and Catalan'.
\newblock In: {\em Proceedings of the Computational Linguistics and Intelligent
  Text Processing 9th International Conference, CICLing 2008}.
\newblock pp. 206--218.

\bibitem[\protect\citeauthoryear{\'Swidzi\'nski}{1992}]{Swidzinski92}
\'Swidzi\'nski, M.: 1992, {\em {Gramatyka formalna j\k{e}zyka polskiego}}.
\newblock Warszawa: Wydawnictwa Uniwersytetu Warszawskiego.

\bibitem[\protect\citeauthoryear{\'Swidzi\'nski}{1994}]{Swidzinski94}
\'Swidzi\'nski, M.: 1994, `Syntactic Dictionary of {Polish} Verbs'.
\newblock Warszawa: Uniwersytet Warszawski / Amsterdam: Universiteit van
  Amsterdam.

\bibitem[\protect\citeauthoryear{Tokarski}{1993}]{Tokarski93}
Tokarski, J.: 1993, {\em {Schematyczny indeks a tergo polskich form
  wyrazowych}}.
\newblock Warszawa: Wydawnictwo Naukowe PWN.

\bibitem[\protect\citeauthoryear{Vapnik}{1995}]{Vapnik95}
Vapnik, V.~N.: 1995, {\em The Nature of Statistical Learning Theory}.
\newblock New York: Springer.

\bibitem[\protect\citeauthoryear{Woli\'{n}ski}{2004}]{Wolinski04}
Woli\'{n}ski, M.: 2004, `{Komputerowa weryfikacja gramatyki
  \'Swidzi\'nskiego}'.
\newblock Ph.D. thesis, Instytut Podstaw Informatyki PAN, Warszawa.

\bibitem[\protect\citeauthoryear{Woli\'{n}ski}{2005}]{Wolinski05}
Woli\'{n}ski, M.: 2005, `An efficient implementation of a large grammar of
  {Polish}'.
\newblock {\em Archives of Control Sciences} {\bf 15(LI), 3}, 251--258.

\bibitem[\protect\citeauthoryear{Woli\'{n}ski}{2006}]{Wolinski06}
Woli\'{n}ski, M.: 2006, `Morfeusz---a Practical Tool for the Morphological
  Analysis of {Polish}'.
\newblock In: M.~A. K{\l}opotek, S.~T. Wierzcho\'n, and K. Trojanowski (eds.):
  {\em Intelligent Information Processing and Web Mining}.
\newblock New York: Springer, pp. 503--512.

\end{thebibliography}

\end{document}